\documentclass{article}

\usepackage[final]{neurips_2025}

\usepackage[utf8]{inputenc} 
\usepackage[T1]{fontenc}    
\usepackage{hyperref}       
\usepackage{url}            
\usepackage{booktabs}       
\usepackage{amsfonts}       
\usepackage{nicefrac}       
\usepackage{microtype}      
\usepackage{xcolor}         

\usepackage{amssymb}
\usepackage{pifont}
\usepackage{diagbox}
\usepackage{lipsum} 
\usepackage{tikz}
\usepackage{atbegshi} 
\usepackage{amsmath}
\usepackage{enumitem}

\usepackage{xspace}          

\title{ VQToken: Neural Discrete Token Representation Learning for Extreme Token Reduction in Video Large Language Models}

\author{%
  Haichao Zhang\thanks{Corresponding author.} \\
  Northeastern University \\
  Boston, MA \\
  \texttt{zhang.haich@northeastern.edu}
  \And
  Yun Fu \\
  Northeastern University \\
  Boston, MA \\
  \texttt{yunfu@ece.neu.edu}
}

\begin{document}

\maketitle

\vspace{-0.35em}
\begin{center}
\small
\textbf{Project:} \href{https://www.zhanghaichao.xyz/VQToken/}{\textcolor[rgb]{0.00,0.33,0.77}{Homepage}}
\quad$\boldsymbol{\cdot}$\quad
\textbf{Code:} \href{https://github.com/Hai-chao-Zhang/VQToken}{\textcolor[rgb]{0.00,0.33,0.77}{GitHub}}
\quad$\boldsymbol{\cdot}$\quad
\textbf{Model:} \href{https://huggingface.co/haichaozhang/VQ-Token-llava-ov-0.5b}{\textcolor[rgb]{0.00,0.33,0.77}{Hugging Face}}
\end{center}
\vspace{0.25em}

\begin{abstract}
Token-based video representation has emerged as a promising approach for enabling large language models (LLMs) to interpret video content. However, existing token reduction techniques, such as pruning and merging, often disrupt essential positional embeddings and rely on continuous visual tokens sampled from nearby pixels with similar spatial–temporal locations. By removing only a small fraction of tokens, these methods still produce relatively lengthy continuous sequences, which falls short of the extreme compression required to balance computational efficiency and token count in video LLMs.
In this paper, we introduce the novel task of \emph{Extreme Short Token Reduction}, which aims to represent entire videos using a minimal set of discrete tokens.  We propose \textbf{VQToken}, a neural discrete token representation framework that
(i) applies adaptive vector quantization to continuous ViT embeddings to learn a compact codebook and 
(ii) preserves spatial–temporal positions via a token hash function by assigning each grid-level token to its nearest codebook entry.
On the Extreme Short Token Reduction task, our VQToken compresses sequences to just \textbf{0.07\%} of their original length while incurring only a \textbf{0.66\%} drop in accuracy on NextQA-MC benchmark. It also achieves comparable performance on ActNet-QA, Long Video Bench, and VideoMME. 
We further introduce the \emph{Token Information Density} (\textbf{TokDense}) metric and formalize fixed-length and adaptive-length subtasks, achieving state-of-the-art results in both settings. Our approach dramatically lowers theoretical complexity, increases information density, way fewer tokens counts, and enables efficient video large language models in resource-constrained environments.

\end{abstract}

\section{Introduction}
\label{sec:intro}

Recent advances in Vision Language Models (VLMs) have enabled unified zero-shot capabilities across diverse tasks, including visual question answering~\cite{xiao2021nextqa, li2024llavaov}, video-to-text generation~\cite{alayrac2022flamingo}, video segmentation~\cite{xu2023hdvila}, and video understanding~\cite{zellers2022merlot}. Although VLMs excel at aligning visual and linguistic information, their substantial computational cost remains a critical bottleneck—especially for video large language models (vLLMs). Video inputs contain spatial-temporal information distributed across numerous frames, resulting in lengthy token sequences that significantly burden computational resources~\cite{vit_dosovitskiy2021image, qwen2}. Consequently, as vLLMs scale in size~\cite{li2024llavaov, zellers2022merlot}, improving computational efficiency becomes imperative.

Unlike textual inputs, video data require tokenizing pixel batches from each frame and concatenating them into extensive sequences. Transformers process these sequences through attention mechanisms at each layer, incurring a computational complexity of $\mathcal{O}(n^2DL)$. As demonstrated in Table~\ref{table:complexity}, the token sequence length ($n$) is the primary contributor to computational overhead, increasing exponentially as the token count grows. This overhead surpasses the influence of model parameters, layers ($L$), and embedding dimensions ($D$). Reducing token sequence length emerges as a promising solution, broadly applicable to most LLMs in a plug-and-play manner.

\begin{table}[t]
\centering
\scalebox{0.53}{
\begin{tabular}{lccccccccc}
\hline
Method & Token Num.$\downarrow$ & Token Num.\%$\downarrow$ & Throughput$\uparrow$ & FLOPs (T)$\downarrow$ & Run-Time$\downarrow$ & Module Complexity$\downarrow$ & LLM Complexity$\downarrow$ & Accuracy$\uparrow$ & TokDense$\uparrow$ \\
\hline
Baseline (LLaVA-OV) & 11664 & 100\% & 46 & 21.91 & 8.2s & 0 & $\mathcal{O}(n^2 d L)$ & 58.38 & 0.005\\
\hline
Token Pruning ($k=0.9n$) & 1152 & 10\% & 89 & 16.09 & 4.3s & $\mathcal{O}(n^2 d)$ & $\mathcal{O}((n-k)^2 d L)$ & 29.12 & 0.025 \\
ToMe ($k=0.9n$) & 1152 & 10\% & 42 & 11.53 & 9.0s & $\mathcal{O}(n^2 d)$ & $\mathcal{O}((n-k)^2 d L)$ & 35.72  & 0.031\\
VidToMe ($k=0.9n$) & 1152 & 10\% & 40 & 11.49 & 9.4s & $\mathcal{O}(n^2 d)$ & $\mathcal{O}((n-k)^2 d L)$ & 39.64 & 0.034 \\
Interpolating ($k=0.73n$) & 3136 & 27\% & 32 & 13.59 & 11.8s & \boldmath $\mathcal{O}(n d)$ & $\mathcal{O}((n-k)^2 d L)$ & 57.20 & 0.018\\
\hline
Ours-Dynamic ($m$ adaptive) & \textbf{13.08} & \textbf{0.07\%} & 49 & 10.50 & 7.8s & \boldmath $\mathcal{O}((n + m^2)d)$ & \boldmath $\mathcal{O}(m^2 d L)$ & \textbf{57.72} & \textbf{4.412}\\
Ours-Fixed ($m=32$) & 32 & 0.14\% & \textbf{91} & \textbf{10.47} & \textbf{4.2s} & \boldmath $\mathcal{O}((n + m^2)d)$ & \boldmath $\mathcal{O}(m^2 d L)$ & 57.46 & 1.796\\
\hline
\end{tabular}}
\caption{
Comparison of model efficiency in terms of token number, throughput, FLOPs, run-time, accuracy, token information density, and complexity analysis. Note: $n$ is the original token count; $k$ is the number of tokens reduced by traditional methods; $m$ is the compressed token count after our extreme reduction approach, with the relationship $n > k \gg m^2 \gg m$; $d$ denotes token dimensionality; and $L$ represents transformer layer count. Given $m \ll n$, our token reduction module has a complexity of $\mathcal{O}((n + m^2) d) \approx \mathcal{O}(n d)$, significantly reducing LLM complexity to $\mathcal{O}(m^2 d L)$. \textit{Module Complexity} quantifies the computational cost of the token reduction method itself, while \textit{LLM Complexity} reflects the computational reduction within the LLM, benefiting from the token reduction. “TokDense” is Token Information Density (accuracy contributed from per token).
}
\label{table:complexity}
\end{table}

Despite extensive efforts to reduce redundancy in video token sequences, existing methods face three main challenges. First, token pruning approaches~\cite{kim2022tokenprune, liu2024revisiting} remove seemingly redundant tokens but often discard critical information, degrading representation quality. Second, token merging techniques—such as ToMe~\cite{tome}, Vid-ToMe~\cite{Vid-ToMe}, and Token Bilinear Interpolating~\cite{li2024llavaov}—group similar tokens without explicit removal; however, they rely on fixed reduction ratios, which limits flexibility and leaves sequences excessively long for large-scale video data. Third, even after pruning or merging, the remaining tokens remain highly contiguous and similar, resulting in low information density and persistent redundancy that impede further compression.

We attribute these challenges to three key limitations. First, existing methods rely on fixed‐count or fixed‐percentage reduction strategies, which either leave sequences overly long, with redundant tokens, or prune so aggressively that critical information is lost. Second, they lack adaptive, context‐sensitive mechanisms for selecting the most informative tokens in the frames. Third, none leverage vector quantization to cluster tokens into discrete categories, hindering substantial gains in information density through thorough compression.
To address these limitations, we propose \textbf{VQToken}, a vector‐quantized token representation framework that dynamically clusters continuous ViT embeddings into a compact, discrete codebook. By mapping each token to its nearest codebook entry, VQToken produces a minimal set of discrete tokens while preserving spatial–temporal relationships. Accurately capturing spatial–temporal dynamics within this discrete clustering, however, remains a critical challenge.

The second major challenge is preserving spatial–temporal coherence during token reduction. Traditional pruning methods~\cite{kim2022tokenprune, liu2024revisiting} often discard positional cues that are vital for tracking object motion accurately. Likewise, similarity‐based merging techniques~\cite{tome, Vid-ToMe, li2024llavaov} tend to ignore spatial–temporal encodings or reapply them inconsistently, which undermines dynamic context modeling.
To overcome this, we introduce a token‐hashing mechanism based on vector quantization. First, each grid‐level token is mapped to its nearest codebook centroid; then, we record its original \((f,h,w)\) index in a three‐dimensional hash table. This table integrates seamlessly into the VQToken architecture, preserving positional information in a compact discrete form. By leveraging the inherent redundancy of video data. Inspired by computationally expensive motion tracking techniques like optical flow, our token‐hashing mechanism offers a lightweight alternative that preserves essential spatial–temporal context with minimal computational overhead.

The third challenge is devising an evaluation framework for highly compressed token sequences. Existing methods neither achieve substantial reduction nor measure information density, making it difficult to compare token–performance trade-offs or assess adaptability. To address this, we define the \emph{Extreme Token Reduction} task with two subtasks: fixed-length compression, which measures LLM accuracy under a predetermined token budget; and adaptive-length compression, which assesses performance when the token count is dynamically determined by video content. We introduce \emph{Token Information Density} (\textbf{TokDense}), defined as accuracy per retained token, to quantify each token’s contribution to task performance. Additionally, we propose separate complexity metrics—one for the token-reduction module itself and another for its impact on downstream LLM inference forming a comprehensive evaluation suite for extreme token reduction methods.

Our contributions are summarized as follows,
\begin{enumerate}
  \item  We present a neural discrete token representation framework, VQToken, that applies adaptive vector quantization to continuous ViT embeddings to learn a compact codebook, and preserves spatial–temporal positions via a hash token function. To the best of our knowledge, this is the first work to leverage vector quantization for token reduction in video large language models.
  \item  A formal definition of the \emph{Extreme Token Reduction} task, together with the \emph{Token Information Density} (\emph{TokDense}) metric and separate complexity measures for the reduction module and downstream LLM inference, covering both fixed‐length and adaptive‐length settings.
  \item  Empirical evidence that VQToken compresses video token sequences to just 0.07\% of their original length with only a 0.66\% drop in NextQA‐MC accuracy, achieving leading efficiency and information density while maintaining competitive performance across multiple benchmarks.
\end{enumerate}

\section{Related Works}

\subsection{Video Large Language Models}
Video Large Language Models (vLLMs) have emerged as powerful tools for bridging video understanding and natural language processing, enabling complex interpretations of video content through language-based interactions. Recent advancements have demonstrated remarkable capabilities in aligning visual and linguistic modalities, exemplified by frameworks such as LLaVA~\cite{liu2023llava, li2024llavaov, liu2023improvedllava, liu2024llavanext}, Flamingo~\cite{alayrac2022flamingo}, AuroraCap~\cite{chai2024auroracap}, VideoOrion~\cite{feng2024videoorion}, and MERLOT Reserve~\cite{zellers2022merlot}. These methods typically rely on extensive pre-training using large-scale datasets like HD-VILA~\cite{xu2023hdvila}, InternVid~\cite{li2023internvid}, and NextQA~\cite{xiao2021nextqa}, generating lengthy token sequences to represent videos effectively.

\subsection{Token Reduction}
Token reduction techniques have gained increasing attention as a means to enhance computational efficiency in Vision Transformers (ViTs)~\cite{vit_dosovitskiy2021image}. Early influential methods, such as Token Pruning~\cite{kim2022tokenprune} and Token Merging (ToMe)~\cite{tome}, significantly reduce computational load by identifying and removing redundant tokens or merging similar tokens, respectively. 
More recently, Video Token Merging (Vid-ToMe)~\cite{Vid-ToMe} has extended ToMe to video inputs, considering temporal redundancy by merging tokens across consecutive frames. Despite these advancements, existing token reduction strategies generally adopt fixed token reduction ratios (e.g., 50\%), limiting flexibility and adaptability. Such fixed strategies can either inadequately reduce redundant tokens, resulting in lingering inefficiencies, or inadvertently merge tokens representing distinct objects, thus losing crucial spatial-temporal dynamics necessary for precise video interpretation. 
To overcome these limitations, our proposed VQ-Token framework introduces dynamic token clustering to generate a compact token representation while explicitly preserving spatial-temporal motion information via a dedicated token indices. Through our novel VQ-Attention mechanism, our approach effectively integrates spatial-temporal coherence into concise token sequences without compromising accuracy, outperforming state-of-the-art token reduction methods, even in scenarios of extreme token compression.

\begin{figure}
  \centering
  \includegraphics[width=\textwidth]{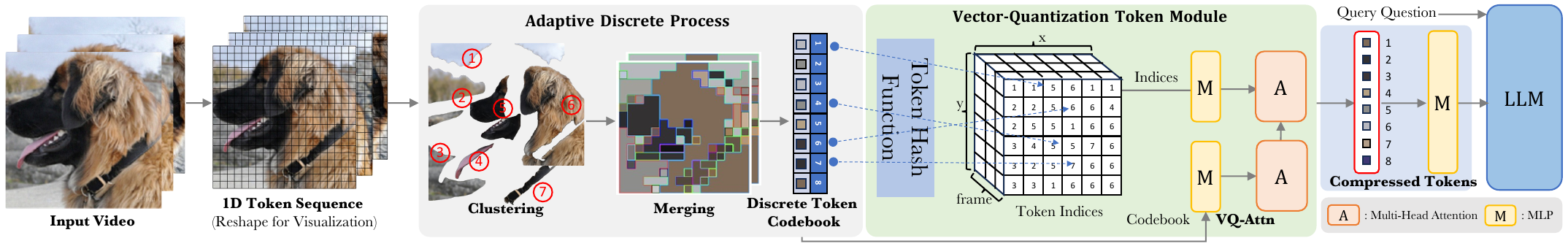}
  \caption{
    \textbf{Overview of neural discrete token representation learning.} 
    First, the input video is tokenized into a continuous sequence of visual tokens. An adaptive discrete process then clusters and vector‐quantizes these tokens into a compact codebook. A token hash function records each token’s original spatial–temporal location and maps it to its nearest codebook entry. The VQ‐Attention module integrates the codebook with the index map to produce a compressed token sequence that preserves positional information. Finally, the compressed tokens and a tokenized query are passed to the large language model for zero‐shot inference.
  }
  \label{fig:overview}
\end{figure}

\section{Methods}
\label{sec:method}

\subsection{Problem Definition: Extreme Token Reduction}

The \emph{Extreme Token Reduction} task aims to compress a long video-derived token sequence \(t\) into a minimal set of tokens without sacrificing downstream performance.

Formally, given a video \(v\) and a query \(q\), a video-language model (vLLM) first tokenizes the video into \(t = \mathrm{Tokenize}(v)\), then uses \(t\) and \(q\) to predict an answer \(a\). A token reduction function \(\mathcal{R}\) maps
\begin{equation}
    t \;\xrightarrow{\;\mathcal{R}\;} t', 
    \quad\text{with}\quad |t'| \ll |t|,
\end{equation}

such that the vLLM’s accuracy on predicting \(a\) remains comparable.

We assess token reduction methods via two subtasks and two complementary complexity metrics:

\textbf{Fixed-Length Reduction.} Each method is evaluated under a predefined token budget \(m\) or reduction ratio \(\rho\), allowing fair comparisons among approaches that require explicit reduction rates.

\textbf{Adaptive-Length Reduction.} Methods dynamically select the optimal \(|t'|\) based on the video’s content complexity, enabling a per-instance trade-off between token count and predictive performance.

Additionally, we introduce two complexity metrics to isolate (i) the computational cost of the reduction module \(\mathcal{R}\), and (ii) the resulting impact on downstream LLM inference.

\subsubsection{Module Complexity and LLM Complexity.}  
Token reduction modules introduce additional computation while reducing the downstream workload of the LLM. To disentangle these effects, we define two complementary metrics:  
\textbf{Module Complexity} measures the computational cost of the token reduction operations alone.  
\textbf{LLM Complexity} quantifies the reduced computational burden on the LLM, reflecting the shorter token sequence length after reduction.

\subsubsection{Token Information Density (TokDense).}  
As token sequences become extremely compact, it is crucial to evaluate how much performance each retained token count contributes. We define TokDense as
\begin{equation}
    \mathrm{TokDense} = \frac{\mathrm{Accuracy}}{\text{Token Count}},
    \label{eq:tokdense}
\end{equation}
where \emph{Accuracy} is measured on the target benchmark and \emph{Token Count} is the number of tokens fed into the LLM after reduction.

\subsection{Neural Discrete Token Representation Learning}

We introduce \emph{Neural Discrete Token Representation Learning}, a vector‐quantization architecture that dynamically minimizes token sequence length while preserving complete spatial–temporal motion information.

\subsubsection{Adaptive Discrete Process}
\label{sec:tok_cluster}

Tokens produced by Vision Transformers (ViTs)~\cite{vit_dosovitskiy2021image} often exhibit temporal continuity and redundancy: continuous visual patterns evolve over time but correspond to discrete semantic entities. Slight variations among tokens can obstruct effective grouping. To address this, we apply vector quantization to cluster similar token embeddings across frames into representative discrete tokens.

Unlike fixed‐ratio merging methods such as ToMe~\cite{tome}, which risk under‐merging or spurious groupings, our adaptive discrete process selects the number of clusters either statically or dynamically. For fixed‐length reduction, we use classical K‐Means~\cite{vassilvitskii2006k}; for adaptive‐length reduction, we employ an adaptive K‐Means variant~\cite{bhatia2004adaptive}. While video‐segmentation approaches (e.g., SAM-based~\cite{ravi2024sam}) can yield fine-grained clusters, their computational overhead makes them less practical for this stage.

Token similarity is measured via cosine similarity. Formally, let \(t_1,\dots,t_N\) denote the original token embeddings and \(K\) the chosen number of clusters. The discrete assignment function \(\mathcal{F}_{\mathrm{disc}}\) produces:
\begin{equation}
\label{eq:cluster}
(s_1, \dots, s_K),\; (c_1, \dots, c_N)
\;=\;\mathcal{F}_{\mathrm{disc}}(t_1, \dots, t_N),
\end{equation}
where \(c_i \in \{1,\dots,K\}\) is the cluster index for token \(t_i\), and \(s_k = \{\,i \mid c_i = k\}\) is the set of token indices assigned to cluster \(k\). This clustering yields a compact discrete codebook of \(K\) representative tokens for subsequent processing.

\subsubsection{Vector‐Quantization Architecture}

To transform the discrete clusters into a compact token sequence while preserving spatial–temporal information, we design three components: a concise codebook, a token hash function, and a VQ‐based reduction module.

\paragraph{Concise Token Codebook.}
Given the original token embeddings \(t_1, \dots, t_N \in \mathbb{R}^D\) and their cluster assignments \(s_1, \dots, s_K\) from Eq.~\ref{eq:cluster}, we build a discrete codebook \(B \in \mathbb{R}^{K \times D}\). Each codebook entry \(b_k\) is computed as the centroid of the embeddings in cluster \(s_k\):
\begin{equation}
    b_k = \frac{1}{|s_k|} \sum_{i \in s_k} t_i,
    \quad
    k = 1, \dots, K.
\end{equation}
Here, \(b_k\) serves as a compact representative for all tokens in cluster \(k\). This codebook captures representative visual patterns and object parts with minimal redundancy.

\paragraph{Token‐Hash Fuction Mapping.}
To retain each token’s original spatial–temporal location, we build a 3D index map \(M \in \{1,\dots,K\}^{T \times H \times W}\). For frame \(f\) and spatial coordinates \((h,w)\), let \(i = f \times (H \cdot W) + h \times W + w\). Then
\begin{equation}
    M_{f,h,w} = c_i,
\end{equation}
where \(T,H,W\) are frame count, height, and width of the ViT grid, and \(c_i\) is the cluster index of token \(t_i\). This mapping preserves positional encodings by recording, for each grid cell, which codebook entry it belongs to.

\paragraph{VQ‐Based Reduction Module.}
We integrate the codebook \(B\) and index map \(M\) via a lightweight VQ‐Attention mechanism using a lightweight VQ‐Attention block that enriches each centroid with motion context without increasing token count,:
\begin{equation}
    \widetilde{M} = \mathrm{MLP}\big(\mathrm{Flatten}(M)\big) \in \mathbb{R}^{K \times D},
\end{equation}
\begin{equation}
    B' = \mathrm{MultiHeadAttn}\big(Q = B W_Q,\; K = B W_K,\; V = \widetilde{M} W_V\big),
\end{equation}
where \(W_Q, W_K \in \mathbb{R}^{D \times D}\) and \(W_V \in \mathbb{R}^{D \times D}\) are learnable projections. The output \(B' \in \mathbb{R}^{K \times D}\) enriches each codebook vector with motion context, yielding the final compressed token set. These tokens are then fed into the downstream vLLM for inference.

\section{Experiments}

\subsection{Implementation Details}

\subsubsection{Training Dataset.}  
We follow the LLaVA-OneVision~\cite{li2024llavaov} setup, using 178K videos from LLaVA-Video-178K dataset paired with 1.3M instruction-following samples (178K captions, 960K open-ended questions, 196K multiple-choice questions) to cover diverse video scenarios.

\subsubsection{Evaluation Benchmarks.}  
\label{sec:benchmarks}
We evaluate on six diverse benchmarks: \textbf{ActivityNet-QA}~\cite{yu2019activitynet} (up to 120 s) for spatio-temporal reasoning on short videos; \textbf{VideoMME}~\cite{fu2024video} (avg. 17 min) for long-video comprehension; \textbf{NExT-QA}~\cite{xiao2021nextqa} for descriptive, causal, and temporal reasoning; \textbf{LongVideoBench}~\cite{wu2025longvideobench} (up to 1 h) for extended narrative understanding; and \textbf{MVBench}~\cite{li2024mvbench} (35 s) comprising 20 reasoning-intensive tasks. These benchmarks collectively test our approach across varied durations, resolutions, and reasoning challenges.

\subsubsection{Training Setup.}  
Starting from the 0.5B-parameter LLaVA-OneVision model~\cite{li2024llavaov} (QWen2 backbone~\cite{qwen2}), we integrate our VQToken framework and fine-tune for zero-shot evaluation. Training uses four NVIDIA A100 GPUs for 85,000 iterations with AdamW and a cosine decay schedule (initial learning rates of \(1\times10^{-5}\) for VQ-Attention and \(2\times10^{-6}\) for the ViT backbone). We employ Zero2 optimization~\cite{rajbhandari2020zero} with batch size 8 and gradient accumulation over 2 steps.

\subsubsection{Metrics.} We report \emph{Accuracy (Acc.)}, the percentage of correct responses on multiple‐choice and open‐ended QA tasks; \emph{Token Count}, the number of tokens processed per example; \emph{Throughput}, measured in frames per second; \emph{FLOPs (T)}, the total tera‐FLOPs required for inference; and \emph{Run‐Time}, the end‐to‐end inference latency. To disentangle the cost of token reduction from its downstream benefit, we also measure \emph{Module Complexity}, the time complexity of the reduction module alone, and \emph{LLM Complexity}, the complexity of the large language model given the reduced token sequence. Finally, \emph{Token Information Density (TokDense)}—defined in Eq.~\ref{eq:tokdense}—quantifies accuracy per retained token.

\subsubsection{Video Large Language Model Baselines.} Due to limited computational resources, we evaluate on the 0.5B‐parameter track, using LLaVA‐OV‐SI and LLaVA‐OneVision as baselines. For efficiency, we integrate our VQToken framework with LLaVA‐OneVision to minimize GPU usage. We also compare against 7B‐parameter versions; despite having 14× more parameters and greater compute, our 0.5B model—using only 0.14\% of the original token count—outperforms some 7B models, highlighting our method’s efficiency.



\subsubsection{Token Reduction Baselines.}
For token reduction, we compare against several baselines:
\textbf{Token Pruning}~\cite{kim2022tokenprune}: A widely recognized method for reducing token numbers and increasing throughput in LLMs.
\textbf{Token Merging (ToMe)}~\cite{tome}: A popular baseline for token reduction, known for its efficiency improvements.
\textbf{Video Token Merging}~\cite{Vid-ToMe}: The current state-of-the-art method for token reduction in video large language models, extending the capabilities of ToMe to video data.
\textbf{Interpolation}: Introduced by~\cite{li2024llavaov}, the use of bilinear interpolation to reduce the number of tokens in visual representations, particularly for video frames. This approach allows the model to handle a larger number of frames by reducing the tokens per frame, achieving a balance between performance and computational cost. (v) \emph{DyCoke}~\cite{tao2024dycoke}, the current SOTA method that employs temporal compression to merge redundant tokens across frames and dynamic KV‐cache reduction to selectively prune spatial redundancy.

\subsection{Quantitative Comparison with vLLM Baselines}

Table~\ref{tab:baseline_comparison} compares our VQ-Token model against recent video–language models introduced between 2022 and 2024. To ensure a fair comparison, we group baselines by model size—0.5B and 7B parameters—accounting for neural scaling effects~\cite{kaplan2020scaling}. Although VQ-Token is trained and evaluated strictly in a zero-shot regime, we also report non-zero-shot baselines fine-tuned on the evaluation datasets for completeness. We evaluate all models using two metrics: accuracy and token count.
Our VQToken slightly outperforms the LLaVA-OneVision baseline in accuracy while reducing the token count from 23,328 (100 \%) to just 32 (0.14 \%), dramatically lowering computational cost. Despite its smaller size (0.5B parameters), VQToken also surpasses several 7B vLLMs in zero-shot accuracy, demonstrating that extreme token reduction can preserve or even improve performance.
These results underscore the effectiveness of our framework in removing redundancy while maintaining essential spatial–temporal and semantic information. The extreme compression achieved by VQ-Token highlights its potential to make large-scale video–language understanding significantly more computationally feasible.

\subsection{Extreme Token Reduction Task}

\subsubsection{Fixed-Length Subtask.}  
To evaluate efficiency under extreme compression, we compare VQ-Token against classical and state‐of‐the‐art reduction methods—Token Pruning, ToMe, and VidToMe—using fixed token budgets. We configure our model with a predetermined cluster size \(K\) and set each baseline to retain exactly the same number of tokens (e.g., 12, 32, or 64). This controlled setting isolates the effect of each reduction strategy on accuracy. As Table~\ref{tab:fixe_comparison} shows, VQ-Token consistently outperforms frame‐level merging (ToMe), sequence‐level merging (VidToMe), and pruning across all extreme budgets, demonstrating superior preservation of semantic content under severe token constraints.

\subsubsection{Adaptive-Length Subtask.}  
Here, each method dynamically selects the optimal token count based on video content complexity. We report both accuracy and the average tokens used per sample (“Avg Tokens”). Table~\ref{tab:adp_comparison} illustrates that, compared to interpolation‐based downsampling and our own fixed‐length variant, the adaptive VQ-Token model achieves higher accuracy while consuming significantly fewer tokens on average. This result underscores its ability to balance efficiency and performance in a content‐aware manner.

\begin{table*}[tb]
\centering
\begin{minipage}[t]{0.48\textwidth}
  \centering
  \caption{\textbf{Fixed-Length Token Reduction.} We evaluate different token reduction approaches by retaining a fixed number of tokens. Each method is adjusted to the same token budgets for fair comparison.}
  \resizebox{0.78\linewidth}{!}{
    \begin{tabular}{l|ccc}
      \hline
      Preset Token Num.$\downarrow$ & \textbf{12} & \textbf{32} & \textbf{64} \\ 
      \hline
      Token Pruning       & 29.12 & 34.50 & 31.31 \\
      ToMe                & 35.72 & 38.50 & 40.10 \\
      VidToMe             & 39.64 & 45.10 & 46.20 \\
      \textbf{Ours (Fixed)} & \textbf{57.03} & \textbf{57.46} & \textbf{57.10} \\
      \hline
    \end{tabular}
  }
  \label{tab:fixe_comparison}
\end{minipage}
\hfill
\begin{minipage}[t]{0.51\textwidth}
  \centering
  \caption{\textbf{Adaptive-Length Token Reduction.} Models select token lengths dynamically based on input sequences. For baselines, we use their default settings optimized for most cases to ensure a fair comparison. }
  \resizebox{\linewidth}{!}{
    \begin{tabular}{l|ccc}
      \hline
      Baseline & Avg. Tokens$\downarrow$ & Acc.$\uparrow$ & TokDense$\uparrow$ \\
      \hline
      Interpolating~\cite{li2024llavaov} & 3136   & 57.20 & 0.018 \\
            Dycoke ~\cite{tao2024dycoke} & 1662.12 & 57.70 & 0.035 \\
      Ours (Fixed)                      & 32     & 57.46 & 1.796 \\
      \textbf{Ours (Dynamic)}                    & \textbf{13.08} & \textbf{57.72} & \textbf{4.413} \\
      \hline
    \end{tabular}
  }
  \label{tab:adp_comparison}
\end{minipage}
\end{table*}

\subsection{Efficiency Comparison and Analysis}

To quantify practical efficiency gains, we compare VQ-Token against existing token reduction methods under standardized settings. For Token Pruning, ToMe, and VidToMe, we retain 10\% of the original tokens; for Interpolation, we use the default setting that retains 27\%. For our approach, \textbf{Ours-Fixed} uses the optimal fixed token count from Table~\ref{tab:fixe_comparison}, and \textbf{Ours-Dynamic} selects token counts adaptively via K-Means~\cite{bhatia2004adaptive}. We evaluate seven metrics: Token Count, Token Ratio (\%), Throughput (clips/sec), FLOPs (T), Run-Time, Module Complexity (reduction module overhead), and LLM Complexity (downstream cost). Using vanilla LLaVA-OV as the backbone, each method is applied and measured under identical conditions.

As Table~\ref{table:complexity} shows, both \textbf{Ours-Fixed} and \textbf{Ours-Dynamic} achieve superior trade-offs between compression and accuracy, reducing theoretical complexity and run-time more than all baselines without sacrificing performance.

\subsection{Performance on Multiple Benchmarks}

\subsubsection{Evaluation Across Diverse Settings.} To test robustness in real-world scenarios—spanning high resolution, long duration, and multi-step reasoning—we evaluate on all benchmarks listed in Sec.~\ref{sec:benchmarks}. Table~\ref{tab:multibench} reports accuracy alongside token reduction relative to the original sequence. Despite compressing tokens by \textbf{99.86\%}, VQ-Token maintains competitive accuracy across tasks, demonstrating its effectiveness and robustness in preserving essential spatial–temporal and semantic information under extreme compression.

\begin{table}[bt]
\centering
\caption{\textbf{Performance across benchmarks.} Despite compressing tokens by 99.86\%, VQ-Token maintains competitive accuracy on diverse video understanding tasks.}
\scalebox{0.8}{
    \begin{tabular}{l|c|cccc}
        \hline
        \textbf{Benchmark} & \textbf{Token \%} & \textbf{NextQA-MC} & \textbf{ActNet-QA} & \textbf{LongVideoBench} & \textbf{VideoMME}  \\
        \hline
        LLaVA-OV-SI            & 100\%   & 53.6  & 49.0  & 41.9  & 40.4  \\
        LLaVA-OneVision        & 100\%   & 57.2  & 50.5  & 45.8  & 43.5  \\
        \textbf{VQ-Token (Ours)} & 0.14\% & 57.4  & 46.3  & 39.3  & 38.2  \\
        \hline
    \end{tabular}
}

\vspace{-5pt}
\label{tab:multibench}
\end{table}

\subsubsection{Evaluation Across Multiple Subtasks.}  
We further evaluate VQ-Token on 20 subtasks from MVBench, covering pose estimation, navigation, multi-step reasoning, and object interactions in dynamic video scenarios. As illustrated in Fig.~\ref{fig:multitask}, our model achieves competitive results across most subtasks and excels in action recognition and object-interaction tasks, demonstrating its ability to focus on critical motion and relational cues.

\begin{table}[t]
\centering
\caption{Zero-shot performance of video–language models. We report accuracy and token reduction relative to the 0.5B LLaVA-OneVision baseline (23,328 tokens = 100\%). Our VQ-Token achieves competitive accuracy with only a 1.6\% drop while using 0.14\% of the original tokens, outperforming other 0.5B models and several 7B models.}
\resizebox{0.80\textwidth}{!}{
\begin{tabular}{lccc|ccc}
\hline
\textbf{Model} & \textbf{\#Parameters} & \textbf{Year} & \textbf{Zero-Shot} & \textbf{Acc.(\%)} $\uparrow$  & \textbf{Token Num.\%} $\downarrow$ \\
\hline
Mistral~\cite{mistral2023} & 7B & 2023 & \checkmark & 51.1  & 100\% \\
P3D-G~\cite{cherian20222} & 7B & 2022 & \ding{55} & 51.3  & 100\% \\
VFC~\cite{momeni2023verbs} & 7B & 2023 & \checkmark & 51.5 & 100\% \\
LLoVi~\cite{llovi2023} & 7B & 2023 & \checkmark & 54.3  & 100\% \\
MVU~\cite{ren2024timechat} & 7B & 2024 & \checkmark & 55.2  & 100\% \\
ATP~\cite{buch2022revisiting} & 7B & 2022 & \ding{55} & 54.3  & 100\% \\
\hline
LLaVA-OneVision~\cite{li2024llavaov} & 0.5B & 2024 & \checkmark & 57.2 & 100\% \\
LLaVA-OV-SI~\cite{li2024llavaov} & 0.5B & 2024 & \checkmark & 53.6  & 27\% \\
\hline
VQ-Token (Ours) & 0.5B & 2024 & \checkmark & 57.5 & 0.14\% \\
\hline
\end{tabular}
}
\label{tab:baseline_comparison}
\end{table}

\begin{table}[tb]
\centering
\caption{\textbf{Ablation study.} We incrementally add each component of VQToken to the LLaVA-OV baseline: discrete codebook, token hash function, and VQ-Attn. “rand” indicates randomized parameters. Each module contributes to accuracy, compression rate, and token information density.}
\resizebox{0.85\columnwidth}{!}{
    \begin{tabular}{cccc|cccc}
        \hline
        VLM & Codebook & Hash Fn. & VQ-Attn & Acc. ↑ & Tokens ↓ & Reduction ↓ & TokDense ↑\\
        \hline
        \checkmark & —            & —             & —             & 57.2 & 23,328 & 100\%   & 0.002 \\
        \checkmark & \checkmark  & —             & —             & 35.2 & 32     & 0.14\% & 1.100 \\
        \checkmark & \checkmark  & \checkmark    & \textit{rand} & 38.9 & 134    & 0.57\% & 0.290 \\
        \checkmark & \checkmark  & \textit{rand} & \checkmark    & 46.9 & 32     & 0.14\% & 1.466 \\
        \checkmark & \textit{rand} & \checkmark  & \checkmark    & 37.7 & 32     & 0.14\% & 1.178 \\
        \checkmark & \checkmark  & \checkmark    & \checkmark    & 57.5 & 32     & 0.14\% & 1.797 \\
        \hline
    \end{tabular}
}

\label{tab:ablation_study}
\end{table}

\subsection{Ablation Study}

\subsubsection{Quantitative Ablation.}  
We evaluate each component’s contribution by incrementally adding the discrete codebook, token hash function (Indices), and VQ attention to the LLaVA‐OV baseline. As shown in Table~\ref{tab:ablation_study}, the \textit{Base} alone compresses tokens to 0.14\% but incurs a 22.0\% accuracy drop, highlighting the loss of spatial–temporal cues. Incorporating the \textit{Indices} marginally improves accuracy, although the LLM cannot yet leverage motion information effectively. Introducing \textit{Attn} restores accuracy substantially, demonstrating that VQ attention is essential to integrate positional context into the compressed tokens. Randomizing each module’s parameters (denoted “rand”) leads to significant performance degradation, confirming the necessity of properly learned codebook, mapping, and attention for extreme token reduction.

\subsubsection{Visualization of Adaptive Discrete Process.}  
\begin{figure}[tb]
\centering
\begin{minipage}[b]{0.58\textwidth}
\centering
  \includegraphics[height=0.15\textheight,keepaspectratio]{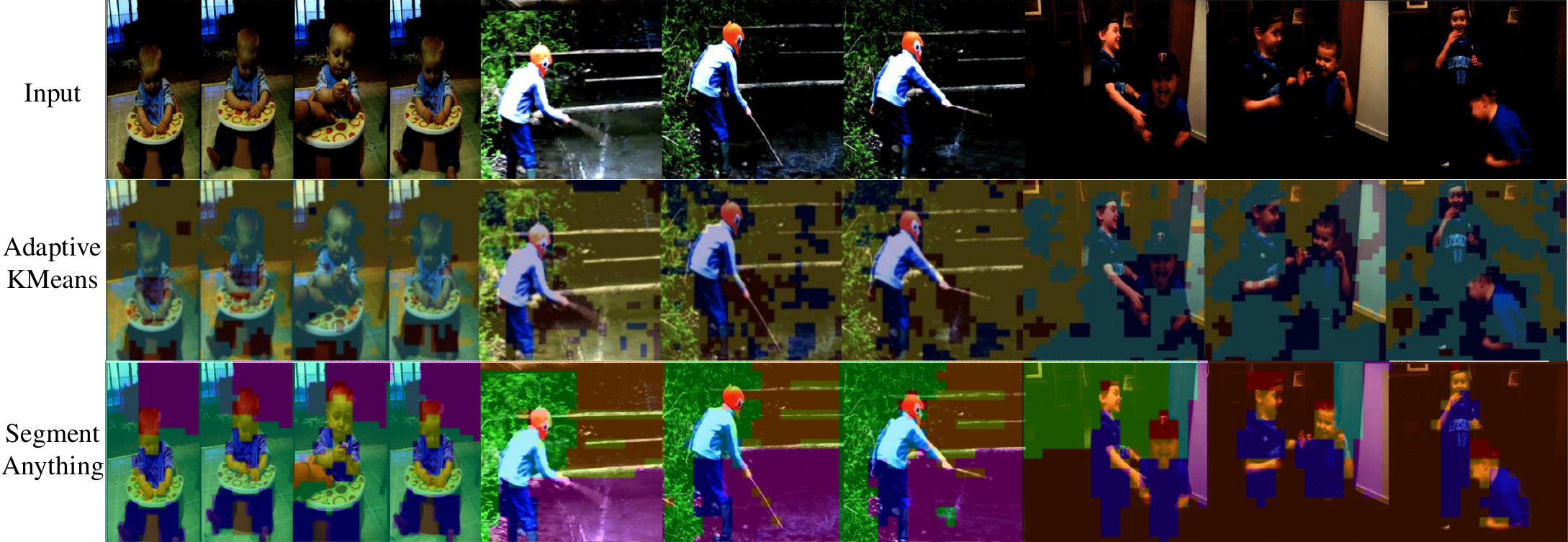}
  \caption{\textbf{Adaptive Discrete Visualization.} Ideally, objects exhibiting similar visual features across frames should be consistently grouped into one or multiple clusters. We compare adaptive K-Means (with reduced cluster numbers for improved visualization) and Segment Anything as adaptive clustering methods.}
  \label{fig:cluster}
\end{minipage}
\hfill
\begin{minipage}[b]{0.40\textwidth}
\centering
  \includegraphics[height=0.2\textheight,keepaspectratio]{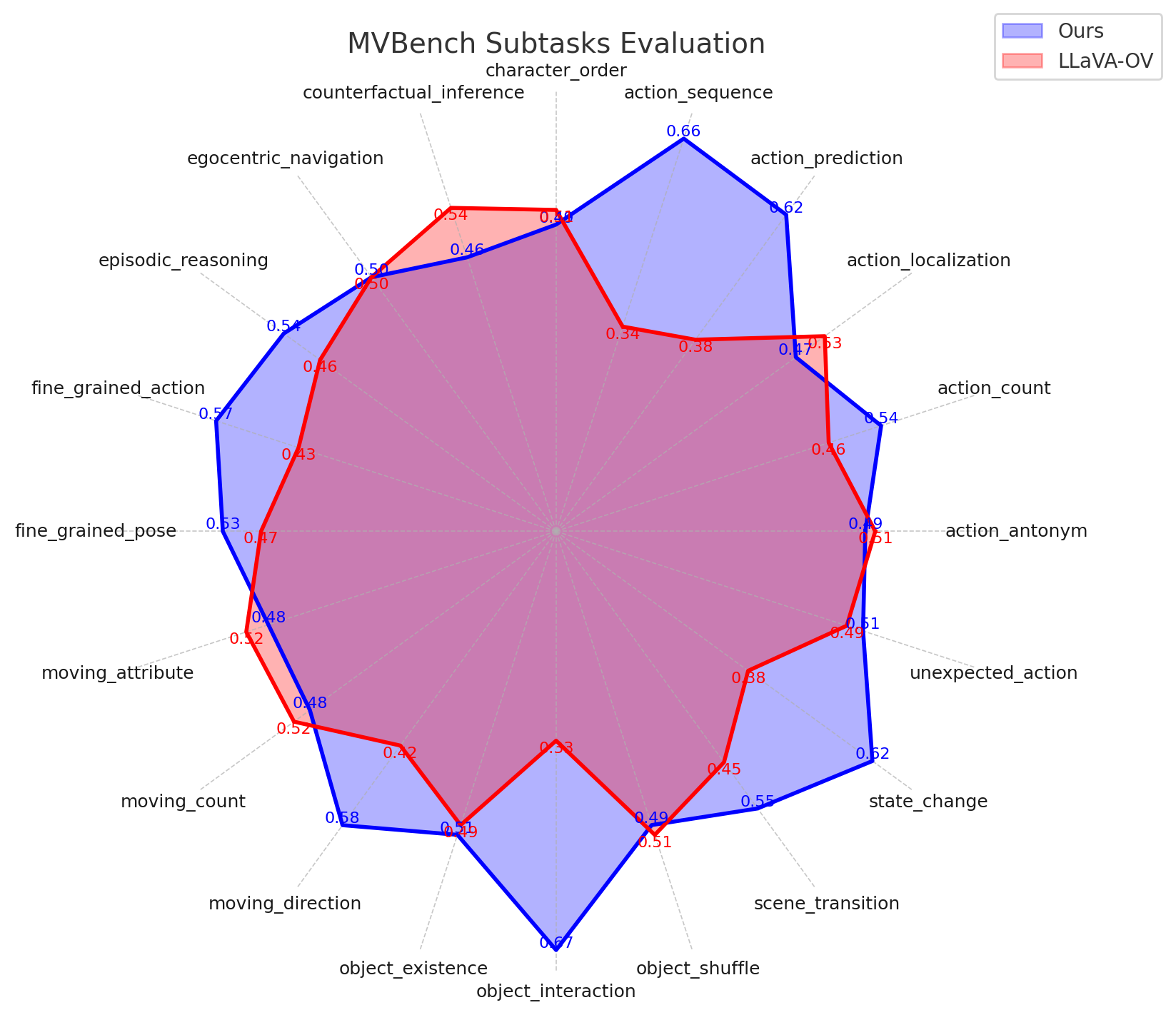}
  \caption{\textbf{MVBench Subtask Performance.} Our method consistently performs well across diverse reasoning and interaction-based tasks, particularly excelling in action recognition and object interaction.}
  \label{fig:multitask}
\end{minipage}
\end{figure}

Figure~\ref{fig:cluster} illustrates clustering behaviors on sample frames. We compare adaptive K‐Means on token embeddings with   the state-of-the-art mask‐based segmentation model Segment Anything (SAM)~\cite{ravi2024sam}. Both methods group semantically similar regions and maintain cluster consistency across frames. While SAM yields finer-grained regions, adaptive K‐Means offers a more computationally efficient alternative that sufficiently captures object trajectories for the token hash function. This efficiency makes adaptive K‐Means the preferred choice for vLLMs requiring extreme compression with minimal overhead.

\section{Conclusion}

We have introduced \textbf{VQToken}, the first neural discrete token representation framework to leverage adaptive vector quantization for extreme token reduction in video large language models. VQToken constructs a compact codebook from continuous ViT embeddings and preserves spatial–temporal positions via a hash‐based token mapping, enabling plug‐and‐play integration with existing architectures.
To benchmark extreme compression, we formalized the \emph{Extreme Token Reduction} task and proposed the \emph{Token Information Density (TokDense)} metric, along with separate complexity measures for the reduction module and downstream LLM inference. These contributions provide a comprehensive evaluation suite for both fixed‐length and adaptive‐length reduction settings.
Empirically, VQToken compresses token sequences to just \textbf{0.07\%} of their original length, achieving over 99.9\% reduction, with only a \textbf{0.66\%} drop in NextQA-MC accuracy. It matches comparable performance on ActivityNet-QA, Long Video Bench, and VideoMME, while delivering state‐of‐the‐art efficiency and information density. Ablation studies confirm that the codebook, token hash function, and VQ‐attention are all critical to preserving semantic and motion information under extreme compression. Efficiency analyses demonstrate substantial reductions in FLOPs, latency, token information density, and overall computational complexity compared to prior methods.
In future work, we will explore hierarchical clustering and learned cluster‐size schedules to further optimize compression, as well as extend the VQToken framework to downstream tasks such as video generation and motion prediction.

{
    \small
    \bibliographystyle{ieeenat_fullname}
    \bibliography{neurips_2025}

\begin{thebibliography}{33}
\providecommand{\natexlab}[1]{#1}
\providecommand{\url}[1]{\texttt{#1}}
\expandafter\ifx\csname urlstyle\endcsname\relax
  \providecommand{\doi}[1]{doi: #1}\else
  \providecommand{\doi}{doi: \begingroup \urlstyle{rm}\Url}\fi

\bibitem[Alayrac et~al.(2022)Alayrac, Donahue, Luc, Miech, Barr, Hasson, Lenc, Mensch, Millican, Reynolds, et~al.]{alayrac2022flamingo}
Jean-Baptiste Alayrac, Jeff Donahue, Pauline Luc, Antoine Miech, Iain Barr, Yana Hasson, Karel Lenc, Arthur Mensch, Katherine Millican, Malcolm Reynolds, et~al.
\newblock Flamingo: a visual language model for few-shot learning.
\newblock \emph{Advances in neural information processing systems}, 35:\penalty0 23716--23736, 2022.

\bibitem[Bhatia et~al.(2004)]{bhatia2004adaptive}
Sanjiv~K Bhatia et~al.
\newblock Adaptive k-means clustering.
\newblock In \emph{FLAIRS}, pages 695--699, 2004.

\bibitem[Bolya et~al.(2023)Bolya, Fu, Dai, Zhang, Feichtenhofer, and Hoffman]{tome}
Daniel Bolya, Cheng-Yang Fu, Xiaoliang Dai, Peizhao Zhang, Christoph Feichtenhofer, and Judy Hoffman.
\newblock Token merging: Your vit but faster.
\newblock In \emph{The Eleventh International Conference on Learning Representations}, 2023.

\bibitem[Buch et~al.(2022)Buch, Eyzaguirre, Gaidon, Wu, Fei-Fei, and Niebles]{buch2022revisiting}
Shyamal Buch, Crist{\'o}bal Eyzaguirre, Adrien Gaidon, Jiajun Wu, Li Fei-Fei, and Juan~Carlos Niebles.
\newblock Revisiting the" video" in video-language understanding.
\newblock In \emph{Proceedings of the IEEE/CVF conference on computer vision and pattern recognition}, pages 2917--2927, 2022.

\bibitem[Chai et~al.(2024)Chai, Song, Du, Meng, Madhavan, Bar-Tal, Hwang, Xie, and Manning]{chai2024auroracap}
Wenhao Chai, Enxin Song, Yilun Du, Chenlin Meng, Vashisht Madhavan, Omer Bar-Tal, Jenq-Neng Hwang, Saining Xie, and Christopher~D Manning.
\newblock Auroracap: Efficient, performant video detailed captioning and a new benchmark.
\newblock \emph{arXiv preprint arXiv:2410.03051}, 2024.

\bibitem[Cherian et~al.(2022)Cherian, Hori, Marks, and Le~Roux]{cherian20222}
Anoop Cherian, Chiori Hori, Tim~K Marks, and Jonathan Le~Roux.
\newblock (2.5+ 1) d spatio-temporal scene graphs for video question answering.
\newblock In \emph{Proceedings of the AAAI Conference on Artificial Intelligence}, pages 444--453, 2022.

\bibitem[Dosovitskiy et~al.(2021)Dosovitskiy, Beyer, Kolesnikov, Weissenborn, Zhai, Unterthiner, Dehghani, Minderer, Heigold, Gelly, Uszkoreit, and Houlsby]{vit_dosovitskiy2021image}
Alexey Dosovitskiy, Lucas Beyer, Alexander Kolesnikov, Dirk Weissenborn, Xiaohua Zhai, Thomas Unterthiner, Mostafa Dehghani, Matthias Minderer, Georg Heigold, Sylvain Gelly, Jakob Uszkoreit, and Neil Houlsby.
\newblock An image is worth 16x16 words: Transformers for image recognition at scale.
\newblock In \emph{International Conference on Learning Representations (ICLR)}, 2021.

\bibitem[Feng et~al.(2024)Feng, Li, Zhang, Luo, Yue, Zheng, and Lu]{feng2024videoorion}
Yicheng Feng, Yijiang Li, Wanpeng Zhang, Hao Luo, Zihao Yue, Sipeng Zheng, and Zongqing Lu.
\newblock Videoorion: Tokenizing object dynamics in videos.
\newblock \emph{arXiv preprint arXiv:2411.16156}, 2024.

\bibitem[Fu et~al.(2024)Fu, Dai, Luo, Li, Ren, Zhang, Wang, Zhou, Shen, Zhang, et~al.]{fu2024video}
Chaoyou Fu, Yuhan Dai, Yongdong Luo, Lei Li, Shuhuai Ren, Renrui Zhang, Zihan Wang, Chenyu Zhou, Yunhang Shen, Mengdan Zhang, et~al.
\newblock Video-mme: The first-ever comprehensive evaluation benchmark of multi-modal llms in video analysis.
\newblock \emph{arXiv preprint arXiv:2405.21075}, 2024.

\bibitem[Jiang et~al.(2023)Jiang, Sablayrolles, Mensch, Bamford, Chaplot, Casas, Bressand, Lengyel, Lample, Saulnier, et~al.]{mistral2023}
Albert~Q Jiang, Alexandre Sablayrolles, Arthur Mensch, Chris Bamford, Devendra~Singh Chaplot, Diego de~las Casas, Florian Bressand, Gianna Lengyel, Guillaume Lample, Lucile Saulnier, et~al.
\newblock Mistral 7b.
\newblock \emph{arXiv preprint arXiv:2310.06825}, 2023.

\bibitem[Kaplan et~al.(2020)Kaplan, McCandlish, Henighan, Brown, Chess, Child, Gray, Radford, Wu, and Amodei]{kaplan2020scaling}
Jared Kaplan, Sam McCandlish, Tom Henighan, Tom~B. Brown, Benjamin Chess, Rewon Child, Scott Gray, Alec Radford, Jeffrey Wu, and Dario Amodei.
\newblock Scaling laws for neural language models.
\newblock \emph{arXiv preprint arXiv:2001.08361}, 2020.

\bibitem[Kim et~al.(2022)Kim, Shen, Thorsley, Gholami, Kwon, Hassoun, and Keutzer]{kim2022tokenprune}
Sehoon Kim, Sheng Shen, David Thorsley, Amir Gholami, Woosuk Kwon, Joseph Hassoun, and Kurt Keutzer.
\newblock Learned token pruning for transformers.
\newblock In \emph{Proceedings of the 28th ACM SIGKDD Conference on Knowledge Discovery and Data Mining}, pages 784--794, 2022.

\bibitem[Lee et~al.(2024)Lee, Wang, Zhang, Fan, and Li]{Vid-ToMe}
Seon-Ho Lee, Jue Wang, Zhikang Zhang, David Fan, and Xinyu Li.
\newblock Video token merging for long-form video understanding.
\newblock In \emph{NIPS}, 2024.

\bibitem[Li et~al.(2024{\natexlab{a}})Li, Zhang, Guo, Zhang, Li, Zhang, Zhang, Li, Liu, and Li]{li2024llavaov}
Bo Li, Yuanhan Zhang, Dong Guo, Renrui Zhang, Feng Li, Hao Zhang, Kaichen Zhang, Yanwei Li, Ziwei Liu, and Chunyuan Li.
\newblock Llava-onevision: Easy visual task transfer.
\newblock \emph{arXiv preprint arXiv:2408.03326}, 2024{\natexlab{a}}.

\bibitem[Li et~al.(2024{\natexlab{b}})Li, Wang, He, Li, Wang, Liu, Wang, Xu, Chen, Luo, et~al.]{li2024mvbench}
Kunchang Li, Yali Wang, Yinan He, Yizhuo Li, Yi Wang, Yi Liu, Zun Wang, Jilan Xu, Guo Chen, Ping Luo, et~al.
\newblock Mvbench: A comprehensive multi-modal video understanding benchmark.
\newblock In \emph{Proceedings of the IEEE/CVF Conference on Computer Vision and Pattern Recognition}, pages 22195--22206, 2024{\natexlab{b}}.

\bibitem[Liu et~al.(2023{\natexlab{a}})Liu, Li, Li, and Lee]{liu2023improvedllava}
Haotian Liu, Chunyuan Li, Yuheng Li, and Yong~Jae Lee.
\newblock Improved baselines with visual instruction tuning, 2023{\natexlab{a}}.

\bibitem[Liu et~al.(2023{\natexlab{b}})Liu, Li, Wu, and Lee]{liu2023llava}
Haotian Liu, Chunyuan Li, Qingyang Wu, and Yong~Jae Lee.
\newblock Visual instruction tuning, 2023{\natexlab{b}}.

\bibitem[Liu et~al.(2024{\natexlab{a}})Liu, Li, Li, Li, Zhang, Shen, and Lee]{liu2024llavanext}
Haotian Liu, Chunyuan Li, Yuheng Li, Bo Li, Yuanhan Zhang, Sheng Shen, and Yong~Jae Lee.
\newblock Llava-next: Improved reasoning, ocr, and world knowledge, 2024{\natexlab{a}}.

\bibitem[Liu et~al.(2024{\natexlab{b}})Liu, Gehrig, Messikommer, Cannici, and Scaramuzza]{liu2024revisiting}
Yifei Liu, Mathias Gehrig, Nico Messikommer, Marco Cannici, and Davide Scaramuzza.
\newblock Revisiting token pruning for object detection and instance segmentation.
\newblock In \emph{Proceedings of the IEEE/CVF Winter Conference on Applications of Computer Vision}, pages 2658--2668, 2024{\natexlab{b}}.

\bibitem[Momeni et~al.(2023)Momeni, Caron, Nagrani, Zisserman, and Schmid]{momeni2023verbs}
Liliane Momeni, Mathilde Caron, Arsha Nagrani, Andrew Zisserman, and Cordelia Schmid.
\newblock Verbs in action: Improving verb understanding in video-language models.
\newblock In \emph{Proceedings of the IEEE/CVF International Conference on Computer Vision}, pages 15579--15591, 2023.

\bibitem[Rajbhandari et~al.(2020)Rajbhandari, Rasley, Ruwase, and He]{rajbhandari2020zero}
Samyam Rajbhandari, Jeff Rasley, Olatunji Ruwase, and Yuxiong He.
\newblock Zero: Memory optimizations toward training trillion parameter models.
\newblock In \emph{SC20: International Conference for High Performance Computing, Networking, Storage and Analysis}, pages 1--16. IEEE, 2020.

\bibitem[Ravi et~al.(2024)Ravi, Gabeur, Hu, Hu, Ryali, Ma, Khedr, R{\"a}dle, Rolland, Gustafson, et~al.]{ravi2024sam}
Nikhila Ravi, Valentin Gabeur, Yuan-Ting Hu, Ronghang Hu, Chaitanya Ryali, Tengyu Ma, Haitham Khedr, Roman R{\"a}dle, Chloe Rolland, Laura Gustafson, et~al.
\newblock Sam 2: Segment anything in images and videos.
\newblock \emph{arXiv preprint arXiv:2408.00714}, 2024.

\bibitem[Ren et~al.(2024)Ren, Yao, Li, Sun, and Hou]{ren2024timechat}
Shuhuai Ren, Linli Yao, Shicheng Li, Xu Sun, and Lu Hou.
\newblock Timechat: A time-sensitive multimodal large language model for long video understanding.
\newblock In \emph{Proceedings of the IEEE/CVF Conference on Computer Vision and Pattern Recognition}, pages 14313--14323, 2024.

\bibitem[Tao et~al.(2024)Tao, Qin, You, Sui, and Wang]{tao2024dycoke}
Keda Tao, Can Qin, Haoxuan You, Yang Sui, and Huan Wang.
\newblock Dycoke: Dynamic compression of tokens for fast video large language models.
\newblock \emph{arXiv preprint arXiv:2411.15024}, 2024.

\bibitem[Vassilvitskii and Arthur(2006)]{vassilvitskii2006k}
Sergei Vassilvitskii and David Arthur.
\newblock k-means++: The advantages of careful seeding.
\newblock In \emph{Proceedings of the eighteenth annual ACM-SIAM symposium on Discrete algorithms}, pages 1027--1035, 2006.

\bibitem[Wang et~al.(2023)Wang, He, Li, Li, Yu, Ma, Li, Chen, Chen, Wang, et~al.]{li2023internvid}
Yi Wang, Yinan He, Yizhuo Li, Kunchang Li, Jiashuo Yu, Xin Ma, Xinhao Li, Guo Chen, Xinyuan Chen, Yaohui Wang, et~al.
\newblock Internvid: A large-scale video-text dataset for multimodal understanding and generation.
\newblock In \emph{The Twelfth International Conference on Learning Representations}, 2023.

\bibitem[Wu et~al.(2025)Wu, Li, Chen, and Li]{wu2025longvideobench}
Haoning Wu, Dongxu Li, Bei Chen, and Junnan Li.
\newblock Longvideobench: A benchmark for long-context interleaved video-language understanding.
\newblock \emph{Advances in Neural Information Processing Systems}, 37:\penalty0 28828--28857, 2025.

\bibitem[Xiao et~al.(2021)Xiao, Shang, Yao, and Chua]{xiao2021nextqa}
Junbin Xiao, Xindi Shang, Angela Yao, and Tat-Seng Chua.
\newblock Next-qa: Next phase of question-answering to explaining temporal actions.
\newblock In \emph{Proceedings of the IEEE/CVF Conference on Computer Vision and Pattern Recognition (CVPR)}, pages 9777--9786, 2021.

\bibitem[Xue et~al.(2022)Xue, Hang, Zeng, Sun, Liu, Yang, Fu, and Guo]{xu2023hdvila}
Hongwei Xue, Tiankai Hang, Yanhong Zeng, Yuchong Sun, Bei Liu, Huan Yang, Jianlong Fu, and Baining Guo.
\newblock Advancing high-resolution video-language representation with large-scale video transcriptions.
\newblock In \emph{International Conference on Computer Vision and Pattern Recognition (CVPR)}, 2022.

\bibitem[Yang et~al.(2024)Yang, Yang, Hui, Zheng, Yu, Zhou, Li, Li, Liu, Huang, Dong, Wei, Lin, Tang, Wang, Yang, Tu, Zhang, Ma, Yang, Xu, Zhou, Bai, He, Lin, Dang, Lu, Chen, Yang, Li, Xue, Ni, Zhang, Wang, Peng, Men, Gao, Lin, Wang, Bai, Tan, Zhu, Li, Liu, Ge, Deng, Zhou, Ren, Zhang, Wei, Ren, Liu, Fan, Yao, Zhang, Wan, Chu, Liu, Cui, Zhang, Guo, and Fan]{qwen2}
An Yang, Baosong Yang, Binyuan Hui, Bo Zheng, Bowen Yu, Chang Zhou, Chengpeng Li, Chengyuan Li, Dayiheng Liu, Fei Huang, Guanting Dong, Haoran Wei, Huan Lin, Jialong Tang, Jialin Wang, Jian Yang, Jianhong Tu, Jianwei Zhang, Jianxin Ma, Jianxin Yang, Jin Xu, Jingren Zhou, Jinze Bai, Jinzheng He, Junyang Lin, Kai Dang, Keming Lu, Keqin Chen, Kexin Yang, Mei Li, Mingfeng Xue, Na Ni, Pei Zhang, Peng Wang, Ru Peng, Rui Men, Ruize Gao, Runji Lin, Shijie Wang, Shuai Bai, Sinan Tan, Tianhang Zhu, Tianhao Li, Tianyu Liu, Wenbin Ge, Xiaodong Deng, Xiaohuan Zhou, Xingzhang Ren, Xinyu Zhang, Xipin Wei, Xuancheng Ren, Xuejing Liu, Yang Fan, Yang Yao, Yichang Zhang, Yu Wan, Yunfei Chu, Yuqiong Liu, Zeyu Cui, Zhenru Zhang, Zhifang Guo, and Zhihao Fan.
\newblock Qwen2 technical report.
\newblock \emph{arXiv preprint arXiv:2407.10671}, 2024.

\bibitem[Yu et~al.(2019)Yu, Xu, Yu, Yu, Zhao, Zhuang, and Tao]{yu2019activitynet}
Zhou Yu, Dejing Xu, Jun Yu, Ting Yu, Zhou Zhao, Yueting Zhuang, and Dacheng Tao.
\newblock Activitynet-qa: A dataset for understanding complex web videos via question answering.
\newblock In \emph{Proceedings of the AAAI Conference on Artificial Intelligence}, pages 9127--9134, 2019.

\bibitem[Zellers et~al.(2022)Zellers, Lu, Lu, Yu, Zhao, Salehi, Kusupati, Hessel, Farhadi, and Choi]{zellers2022merlot}
Rowan Zellers, Jiasen Lu, Ximing Lu, Youngjae Yu, Yanpeng Zhao, Mohammadreza Salehi, Aditya Kusupati, Jack Hessel, Ali Farhadi, and Yejin Choi.
\newblock Merlot reserve: Neural script knowledge through vision and language and sound.
\newblock In \emph{2022 IEEE/CVF Conference on Computer Vision and Pattern Recognition, CVPR 2022}, pages 16354--16366. IEEE Computer Society, 2022.

\bibitem[Zhang et~al.(2023)Zhang, Lu, Islam, Wang, Yu, Bansal, and Bertasius]{llovi2023}
Ce Zhang, Taixi Lu, Md~Mohaiminul Islam, Ziyang Wang, Shoubin Yu, Mohit Bansal, and Gedas Bertasius.
\newblock A simple llm framework for long-range video question-answering, 2023.

\end{thebibliography}
}

\end{document}